# Bengali Fake Review Detection using Semi-supervised Generative Adversarial Networks

Md. Tanvir Rouf Shawon§, G. M. Shahariar§, Faisal Muhammad Shah, Mohammad Shafiul Alam,
and Md. Shahriar Mahbub
Ahsanullah Unviersity of Science and Technology, Dhaka, Bangladesh
Email: {shawontanvir, shahariar_shibli, faisal, shafiul, shahriar}.cse@aust.edu

*Abstract*—This paper investigates the potential of semi-supervised Generative Adversarial Networks (GANs) to fine-tune pretrained language models in order to classify Bengali fake reviews from real reviews with a few annotated data. With the rise of social media and e-commerce, the ability to detect fake or deceptive reviews is becoming increasingly important in order to protect consumers from being misled by false information. Any machine learning model will have trouble identifying a fake review, especially for a low resource language like Bengali. We have demonstrated that the proposed semi-supervised GAN-LM architecture (generative adversarial network on top of a pretrained language model) is a viable solution in classifying Bengali fake reviews as the experimental results suggest that even with only 1024 annotated samples, BanglaBERT with semi-supervised GAN (SSGAN) achieved an accuracy of 83.59% and a f1-score of 84.89% outperforming other pretrained language models - BanglaBERT generator, Bangla BERT Base and Bangla-Electra by almost 3%, 4% and 10% respectively in terms of accuracy. The experiments were conducted on a manually labeled food review dataset consisting of total 6014 real and fake reviews collected from various social media groups. Researchers that are experiencing difficulty recognizing not just fake reviews but other classification issues owing to a lack of labeled data may find a solution in our proposed methodology.

## I. INTRODUCTION

Social media platforms enable anybody to submit unrestricted comments or criticism about any services, items, brands, or businesses at any time. Because there are no constraints, people use social media to unjustly advertise specific services, products, or items or to disparage competitors. Customers decide whether to buy a specific product or change their minds after reading reviews on social media. These reviews provide a first-hand perspective of the reviewer's experience and serve as an excellent resource for any potential customers. For instance, before people decide to visit a restaurant, they read reviews on what previous diners thought of the menu items or service. They choose whether or not to visit the restaurant based on the feedback from the reviews. Reviews are thought of as genuine means of communicating positive or negative opinion about products and services, hence any attempt to falsify review by including false or deceptive information in them is regarded to be unethical. Fake review detection is challenging because reviews can often be written in such a way that they appear genuine, making them hard to differentiate from real reviews. Additionally, fake reviews are often created with the intention of manipulating public opinion, which makes it even more difficult to detect them accurately. Furthermore, language used in reviews can vary greatly depending on the user's background and cultural context which complicates the task of detecting fake reviews. There have been a plenty of studies conducted to create computational models for fake review detection i.e. traditional machine learning based approach [1], deep learning based approach [2], GAN based approach [3] and transformer based approach [4]. The majority of recent research has been on supervised learning techniques, which need labeled data and are scarce when it comes to identifying fake online reviews because it is exceedingly difficult, if not impossible, to accurately label fraudulent reviews manually. While there are several research works available for languages like English, Persian, and Roman-Urdu, there are none for Bengali. In order to identify fake reviews using a dataset of manually obtained and annotated Bengali fake review texts, we have therefore proposed a semi-supervised generative adversarial network architecture with five pretrained language models in this study.

## II. RELATED WORKS

Semi-supervised GAN-BERT architecture proposed by Croce et al. [5] was first employed to fine-tune pretrained BERT on Bengali language by Raihan et al. [6] in order to categorize Bengali texts with a few labeled examples. They evaluated how well GAN-Bangla-BERT performed on two downstream Bengali tasks (hate speech and fake news detection) in comparison to Bangla-Electra and Bangla BERT Base. Semi-supervised GAN-BERT architecture with task and language specific pretrained language model as encoder was employed in several downstream tasks. Using GAN-BERT, Ta et al. [7] identified violent and abusive social media posts in Spanish. However, noise vectors were tweaked using random rate different from the original SS-GAN architecture before being fed to the generator network. Zaharia et al. [8] utilized Romanian BERT with SS-GAN for Romanian dialect identification while Yusuf et al. [9] fine-tuned ARBERT and MARBERT with SS-GAN for Arabic dialect identification. Colón-Ruiz and Segura-Bedmar [10] employed a BERT model followed by

§Authors have equal contributions.

Bidirectional Long Short Term Memory (Bi- LSTM) network with SS-GAN for sentiment analysis on drug reviews. Auti et al. [11] utilized BioBERT with SS-GAN which performed best to classify pharmaceutical compliant and non-compliant texts. Recently, Athirai et al. [12] has proposed an adversarial training strategy for recognizing fake review with a few labeled data and a large number of unlabeled data using Generative Pre-trained Transformer 2 (GPT-2) that can also generate synthetic fake and non-fake reviews with appropriate perplexity, resulting in more labeled data for training.

## III. BACKGROUND STUDY

### A. Semi-supervised Generative Adversarial Networks

Generative modeling is an unsupervised learning problem also referred to as "adversarial learning" where a model automatically identifies and learns the pattern or distribution from the input data and may generate new synthetic instances that are identical to the original data. Generative Adversarial Network (GAN) [13] is an architecture for training generative models. Semi-supervised GANs (SS-GANs) [14] are enhancement of the GAN framework where both labeled and unlabeled data are utilized. The training process of GAN involves a generator model and a discriminator model. The generator model generates a batch of samples (fake). The generated and original samples are then provided to the discriminator model which identifies the samples as either real (original data) or fake (generated data). In SS-GAN, the discriminator not only determines if a sample was generated or not, but also assigns a target category to each sample. The generator and discriminator models are trained in an adversarial setting, but the discriminator is trained over a (n + 1) class objective. While performing a binary representation classification (real or fake) of samples into the $(n+1)^{th}$ class, the discriminator also categorizes real samples into one of the (1, ..., n) classes. The discriminator is trained with labeled data, while the unlabeled input data along with the generated samples from the generator are leveraged to help in learning the representation of the data. The higher the number of unlabeled data is, the better for the model to capture the representation. Although SS-GANs are commonly used with computer vision, Croce et al. [5] showed that they can also be used in fine-tuning language model on textual features referred to as "GAN-BERT". On top of BERT [15], the authors employed the SS-GAN architecture which creates the opportunity to fine-tune any pretrained language model using as the encoder in the SS-GAN architecture, hence in this study we refer the architecture as "GAN-LM".

### B. Pretrained Language Models on Bengali

Though building language-specific models is challenging for languages with low resources such as Bengali, but recently, a few pretrained language models have gained prominence by providing state-of-the-art (SOTA) results on a variety of downstream tasks which are briefly discussed below.

**1) BERT based**: Bidirectional Encoder Representations from Transformers (BERT) [15] is built on a deep learning model in which each input and output element is linked, and the weights between them are dynamically determined depending on that relationship. The novelty of BERT is its capability of bidirectional training where instead of only learning the word that comes before or after another, the language model learns the context of the word depending on its surroundings. **Bangla BERT Base**[1] is a pretrained Bengali language model based on mask language modeling, as defined in BERT.

TABLE I: Configurations of Bengali language models

| Pretrained Language Models | Type | Params | Embedding Size |
|---|---|---|---|
| Bangla BERT Base | BERT-base | 110M | 768 |
| BanglaBERT | ELECTRA-base | 110M | 768 |
| BanglaBERT generator | ELECTRA-base | 34M | 768 |
| sahajBERT | ALBERT-large | 18M | 128 |
| Bangla-Electra | ELECTRA-small | 14M | 128 |

**2) ELECTRA based**: The pre-training task for Efficiently Learning an Encoder that Accurately Classifies Token Replacements (ELECTRA) [16] is based on identifying replaced tokens in the input sequence. A discriminator model is taught to recognize which tokens have been replaced in a corrupted sequence, while a generator model is trained to predict the original tokens for masked out tokens. The setup is comparable to a GAN training system, with the exception that it is not adversarial because the generator is not trained to try to trick the discriminator. **Bangla-Electra**[2] is an ELECTRA model pretrained with the replacement token detection (RTD) objective, **BanglaBERT** [17] is an ELECTRA discriminator model, and **BanglaBERT generator** [17] is an ELECTRA generator model pretrained with the masked language modeling (MLM) goal on sizable quantities of Bengali corpora

**3) ALBERT based**: A Lite BERT (ALBERT) [18] demonstrated that better language models do not always imply larger models by using the same encoder segment architecture from the original Transformer with three key differences: factorized embedding parameters, cross-layer parameter sharing, and Sentence-order prediction (SOP) instead of Next Sentence Prediction (NSP). **sahajBERT**[3] is a collaboratively pretrained ALBERT model on Bengali language using masked language modeling (MLM) and Sentence Order Prediction (SOP) objectives. Table I represents the various configurations of the Bengali pretrained language models utilized in this study.

## IV. PROPOSED METHODOLOGY

The proposed approach of this study is divided into four steps discussed below. The main difference of the proposed workflow with [5] and [6] is not much in the architecture rather in the usage of the language specific BERT models. The methodology described in [6] only used Bangla BERT Base in their architecture whereas we utilized several widely available Bengali pretrained language models such as BanglaBERT, BanglaBERT generator, Bangla-Electra and sahajBERT.

---

[1] https://github.com/sagorbrur/bangla-bert
[2] https://huggingface.co/monsoon-nlp/bangla-electra
[3] https://huggingface.co/neuropark/sahajBERT

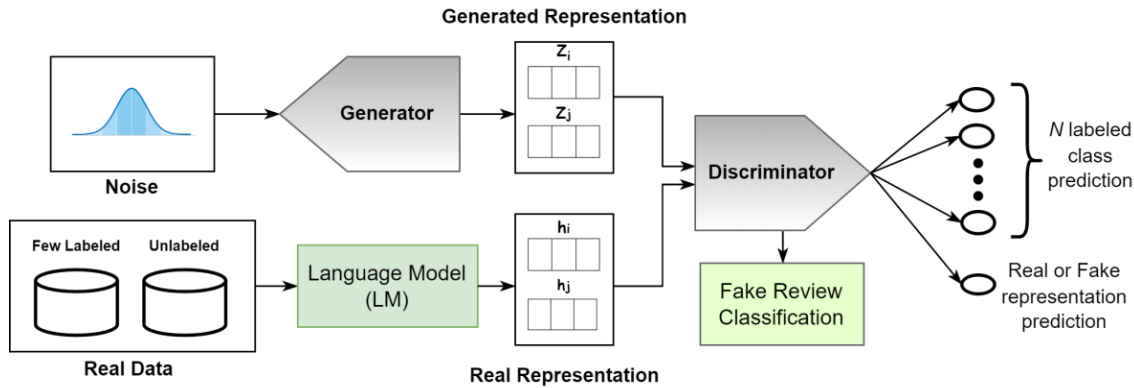

Fig. 1: Proposed methodology for fine-tuning language models with semi-supervised GAN architecture

**Step 1) Text Normalization:** The text normalization module specified in [19] for cleaning Bengali text is employed to preprocess the entire dataset. Text normalization involves managing multiple white spaces, replacing URLs, emojis, and unicode, as well as fixing single or double quotes.

**Step 2) Data Distribution** : The dataset is divided into three parts: labeled, unlabeled and test samples. The data distribution is performed based on the values listed in Table II. In all the experiments, the number of unlabeled samples is 512 while the number of labeled samples is gradually increased.

**Step 3) Fine-tuning LMs with Semi-supervised GAN** : The semi-supervised GAN-LM architecture presented in Fig 1 is similar to the architecture of Croce et al. [5] with a generator and a discriminator which are multi-layer perceptrons (MLPs) employed in this step along with the pretrained language models available in Bengali as encoders discussed in section III-B. The generator model takes input a 100 dimension noise vector that follows Gaussian distribution and outputs a representation vector of dimension equal to the embedding size of the language model. The labeled and unlabeled samples from the previous step are provided to the language model as input and the output is the sentence embeddings i.e. real distribution vectors of dimension equal to the generated representation.

TABLE II: Data Distribution for the experimentations

| No. of Labeled Samples | Unlabeled Samples | Testing Samples |
|---|---|---|
| 32 | 512 | 512 |
| 64 | 512 | 512 |
| 128 | 512 | 512 |
| 256 | 512 | 512 |
| 512 | 512 | 512 |
| 1024 | 512 | 128 |

The generated and real distribution vectors are provided as input to the discriminator. The discriminator learns not only to classify the representations of real instances into one of the (1, ..., n) classes but also to identify whether the representation is real or fake by setting values 0 or 1 into the $(n+1)^{th}$ class. For the semi-supervised GAN-LM architecture used in this research, we adopted the same adversarial training objectives and loss functions as those specified in [5]. The weights of the language model are adjusted during back propagation together with the discriminator to improve the intermediate representations of the labeled and unlabeled data. The pretrained language model is used for inference after training while the generator is removed.

**Step 4) Evaluation using performance metrics** : The test samples are used to measure the performance of the fine-tuned language model. For performance measurement and comparison, evaluation metrics such as accuracy, precision, recall and F1-score are used.

## V. EXPERIMENTAL RESULTS & EVALUATIONS

### A. Dataset Description

The dataset used in this study comprises of total 6014 fake and authentic reviews written in Bengali language and collected from selected publicly accessible Facebook groups where Bengali speakers provide reviews on food items on a regular basis. We manually gathered the posts made by the members of the groups regarding the food items and services of various restaurants and labeled them as authentic or fake. After majority voting on the annotations performed by three different individuals, we gathered 871 fake and 5015 authentic reviews respectively.

### B. Hyperparameter Setting

For all the experiments, we used a batch size of 16, a learning rate of 5e-5 for both the discriminator and generator, AdamW as optimizer, Softmax as activation function, and number of training epochs for BanglaBERT, BanglaBERT Generator, Bangla BERT Base, Bangla-Electra, and sahajBERT is 7, 18, 25, 13, and 26 respectively.

### C. Experimental Result and Analysis

We fine-tuned BanglaBERT with 1024 labeled samples and found an accuracy of 65.62% which is very low for a binary classification task which motivated us to use the GAN-LM architecture. The performance comparison between fine-tuned model and five different Bengali language models with semi-supervised GAN architecture is presented in Table III and a graphical representation of test accuracy with respect to number of epochs is depicted in Fig 2. **BanglaBERT** has outperformed all other language models in GAN-LM architecture for our dataset. From Table III, it is clear that BanglaBERT has achieved the highest 83.59% accuracy which is almost 3% and 4% higher than BanglaBERT generator and Bangla BERT Base respectively and 10% higher than Banla-Electra with 1024 labeled samples. The f1-score of 84.89% supports the model's highest competency towards not giving the wrong prediction in both cases among all the models. It also achieved a good accuracy of 69.53% and 71.09% only with 32 and 64 samples respectively.

TABLE III: Performance comparison of different language models

| Model | No. of Labeled Sample | Accuracy | Precision | Recall | F1 – Score |
|---|---|---|---|---|---|
| Fine Tuned Bangla BERT | 1024 | **0.65625** | 0.68000 | 0.54838 | 0.60714 |
| GAN Bangla BERT | 32 | **0.69531** | 0.71179 | 0.64427 | 0.67635 |
|  | 64 | **0.71093** | 0.76650 | 0.59684 | 0.67111 |
|  | 128 | 0.73047 | 0.73469 | 0.71146 | 0.72289 |
|  | 256 | 0.75391 | 0.71381 | 0.83795 | 0.77091 |
|  | 512 | 0.76563 | 0.78298 | 0.72727 | 0.75410 |
|  | 1024 | **0.83593** | 0.84286 | 0.85507 | **0.84892** |
| GAN Bangla BERT Generator | 32 | **0.65234** | 0.65060 | 0.64032 | 0.64542 |
|  | 64 | **0.67578** | 0.66540 | 0.69170 | 0.67830 |
|  | 128 | 0.71680 | 0.70930 | 0.72332 | 0.71624 |
|  | 256 | 0.77539 | 0.75746 | 0.80237 | 0.77927 |
|  | 512 | 0.78711 | 0.78571 | 0.78261 | 0.78416 |
|  | 1024 | **0.80468** | 0.82353 | 0.81159 | 0.81752 |
| GAN Bangla BERT Base | 32 | 0.55469 | 0.55760 | 0.47826 | 0.51489 |
|  | 64 | 0.64258 | 0.62868 | 0.67589 | 0.65143 |
|  | 128 | 0.67188 | 0.66798 | 0.66798 | 0.66798 |
|  | 256 | 0.68555 | 0.69328 | 0.65217 | 0.67210 |
|  | 512 | 0.73633 | 0.72519 | 0.75099 | 0.73786 |
|  | 1024 | **0.79687** | **0.79452** | **0.84058** | 0.81690 |
| GAN Bangla-Electra | 32 | 0.55273 | 0.55405 | 0.48617 | 0.51790 |
|  | 64 | 0.58398 | 0.60638 | 0.45059 | 0.51701 |
|  | 128 | 0.65234 | 0.64591 | 0.65613 | 0.65098 |
|  | 256 | 0.65234 | 0.61610 | 0.78656 | 0.69097 |
|  | 512 | 0.68359 | 0.64444 | 0.80237 | 0.71479 |
|  | 1024 | **0.73437** | **0.72727** | **0.81159** | 0.76712 |
| GAN sahaj BERT | 32 | 0.51172 | 0.50319 | 0.93676 | 0.65470 |
|  | 64 | **0.68750** | 0.72683 | 0.58893 | 0.65066 |
|  | 128 | 0.66016 | 0.67873 | 0.59289 | 0.63291 |
|  | 256 | **0.72852** | 0.70956 | 0.76285 | 0.73524 |
|  | 512 | 0.73438 | 0.71910 | 0.75889 | 0.73846 |
|  | 1024 | 0.46094 | N/A | N/A | N/A |

Fig 2a shows the stability of BanglaBERT model where the accuracy is continuously increasing for 32 to 1024 labeled samples without much fluctuations like other language models. **BanglaBERT generator** with only 34M parameters achieved the second-highest accuracy among other language models considering 1024 labeled samples. It only trails BanglaBERT by 3.12% but beats other models in terms of accuracy. The accuracy considering 32 and 64 samples are also quite comparable with 65.23% and 67.57%. The fluctuations in the test accuracy of BanglaBERT generator with respect to the number of epochs in Fig 2b clearly exhibit the full capacity of the lower number of parameters of the model. Despite having a higher number of parameters (110M) **Bangla BERT Base** performed almost same as BanglaBERT generator. The model's accuracy is 79.68%, which is extremely close to that of the BanglaBERT generator, but it got lower precision score indicating its' tendency towards making more false positive predictions than the BanglaBERT generator with 1024 labeled samples. Fig 2c shows the instability of the model where a lot of ups and downs in the test accuracy can be seen up to 25 epochs. **Bangla-Electra** has the lowest number of parameters (14M) among all the models but its' performance is quite noticeable in terms of the parameters it holds. Though the model attained an accuracy of 73.43% with 1024 labeled samples but with a precision score of 72.72% and recall score of 81.15%, it demonstrated considerable difficulty in identifying false positive cases. The model hallucinated as it predicted a significant amount of fake data as authentic with increasing number of labeled samples. The model failed to perform when the labeled data is very low. Though the model did not perform as other language models with a few labeled data, but it showed a smooth progress in test accuracy as the number of epochs increased which can be observed in Fig 2d. The performance of **sahajBERT** is also surprising as it gave a reasonable accuracy with a few labeled data. It gave a comparable accuracy of 68.75% and 72.85% considering 64 and 256 labeled data which is very close to other models but failed to perform when the number of labeled examples was increased to 1024. The smaller number of parameters of the model might be responsible for not accurately detecting the test data in case when the difference in representations between fake and authentic reviews in the dataset was low. Fig 2e explains the behavior of the model because it shows how the test accuracy fluctuates with respect to the number of epochs. It performed poor with 32 labeled data, but after 15 training epochs, performed well with 64 to 512 labeled samples, and after 26 epochs, it stabilizes.

VI. CONCLUSION AND FUTURE WORK

Fine-tuning language models with semi-supervised GAN architecture is a viable solution for classifying Bengali fake reviews with a few annotated data. Experimental results show that even with only 1024 annotated samples our proposed method performs better in classifying reviews after fine-tuning BanglaBERT and Bangla BERT Base in a semi-supervised manner. Despite having fewer parameters, the Bangla-Electra, SahajBERT, and BanglaBERT generator also produced decent results. The exploration of how GANs may be utilized to generate Bengali reviews will be an intriguing development.

ACKNOWLEDGMENT

This research work is conducted under " Bengali Fake Reviews: A Benchmark Dataset and Detection System" project funded by AUST Internal Research Grant.

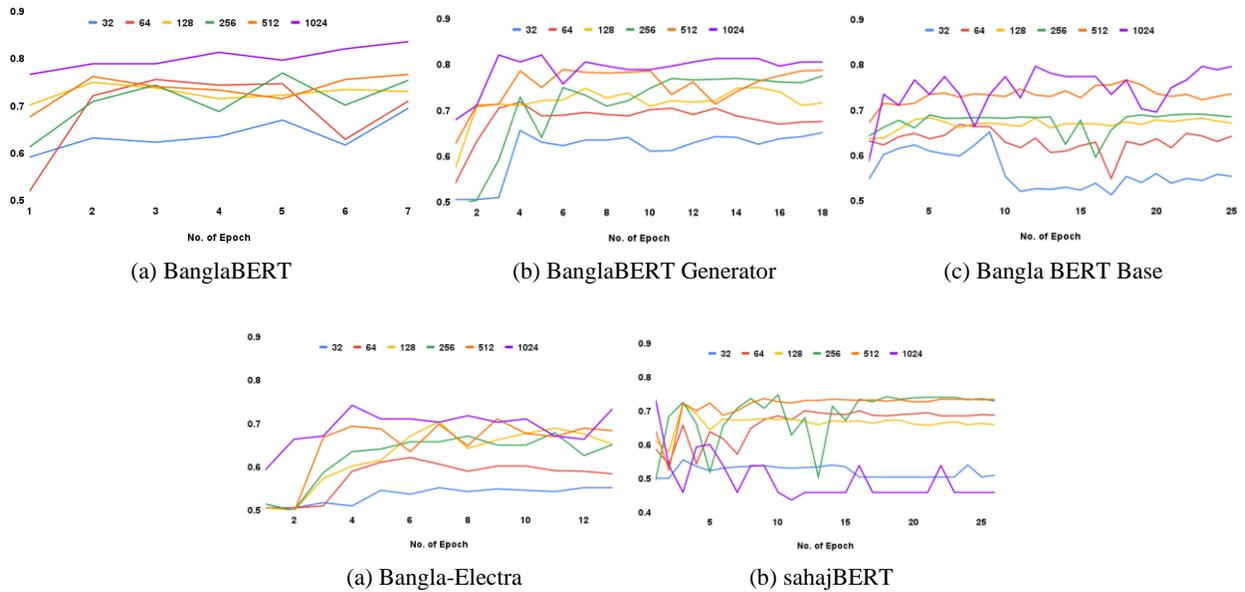

Fig. 2: Test accuracy vs. #epochs of the experimental models with 32, 64, 128, 256, 512, and 1024 labeled samples

## REFERENCES

[1] Jindal, Nitin, and Bing Liu. "Opinion spam and analysis." In Proceedings of the 2008 international conference on web search and data mining, pp. 219-230. 2008.

[2] Shahariar, G. M., Swapnil Biswas, Faiza Omar, Faisal Muhammad Shah, and Samiha Binte Hassan. "Spam review detection using deep learning." In 2019 IEEE 10th Annual Information Technology, Electronics and Mobile Communication Conference (IEMCON), pp. 0027-0033. IEEE, 2019.

[3] Aghakhani, Hojjat, Aravind Machiry, Shirin Nilizadeh, Christopher Kruegel, and Giovanni Vigna. "Detecting deceptive reviews using generative adversarial networks." In 2018 IEEE Security and Privacy Workshops (SPW), pp. 89-95. IEEE, 2018.

[4] Kennedy, Stefan, Niall Walsh, Kirils Sloka, Jennifer Foster, and Andrew McCarren. "Fact or factitious? Contextualized opinion spam detection." arXiv preprint, arXiv:2010.15296 (2020).

[5] Croce, Danilo, Giuseppe Castellucci, and Roberto Basili. "GAN-BERT: Generative adversarial learning for robust text classification with a bunch of labeled examples." In Proceedings of the 58th annual meeting of the association for computational linguistics, pp. 2114-2119. 2020.

[6] Tanvir, Raihan, Md Tanvir Rouf Shawon, Md Humaion Kabir Mehedi, Md Motahar Mahtab, and Annajiat Alim Rasel. "A GAN-BERT Based Approach for Bengali Text Classification with a Few Labeled Examples." In International Symposium on Distributed Computing and Artificial Intelligence, pp. 20-30. Springer, Cham, 2023.

[7] Ta, Hoang Thang, Abu Bakar Siddiqur Rahman, Lotfollah Najjar, and Alexander Gelbukh. "GAN-BERT: Adversarial Learning for Detection of Aggressive and Violent Incidents from Social Media." In Proceedings of the Iberian Languages Evaluation Forum (IberLEF 2022), CEUR Workshop Proceedings. CEUR-WS. org. 2022.

[8] Zaharia, George-Eduard, Andrei-Marius Avram, Dumitru-Clementin Cercel, and Traian Rebedea. "Dialect identification through adversarial learning and knowledge distillation on romanian bert." In Proceedings of the Eighth Workshop on NLP for Similar Languages, Varieties and Dialects, pp. 113-119. 2021.

[9] Yusuf, Mahmoud, Marwan Torki, and Nagwa M. El-Makky. "Arabic Dialect Identification with a Few Labeled Examples Using Generative Adversarial Networks." In Proceedings of the 2nd Conference of the Asia-Pacific Chapter of the Association for Computational Linguistics and the 12th International Joint Conference on Natural Language Processing, pp. 196-204. 2022.

[10] Colón-Ruiz, Cristóbal, and Isabel Segura-Bedmar. "Semi-Supervised Generative Adversarial Network for Sentiment Analysis of drug re- views." (2021).

[11] Auti, Tapan, Rajdeep Sarkar, Bernardo Stearns, Atul Kr Ojha, Arindam Paul, Michaela Comerford, Jay Megaro, John Mariano, Vall Herard, and John Philip McCrae. "Towards Classification of Legal Pharmaceutical Text using GAN-BERT." In Proceedings of the First Computing Social Responsibility Workshop within the 13th Language Resources and Evaluation Conference, pp. 52-57. 2022.

[12] Irissappane, Athirai A., Hanfei Yu, Yankun Shen, Anubha Agrawal, and Gray Stanton. "Leveraging GPT-2 for classifying spam reviews with limited labeled data via adversarial training." arXiv preprint, arXiv:2012.13400 (2020).

[13] Goodfellow, Ian, Jean Pouget-Abadie, Mehdi Mirza, Bing Xu, David Warde-Farley, Sherjil Ozair, Aaron Courville, and Yoshua Bengio. "Generative adversarial networks." Communications of the ACM 63, no. 11 (2020): 139-144.

[14] Salimans, Tim, Ian Goodfellow, Wojciech Zaremba, Vicki Cheung, Alec Radford, and Xi Chen. "Improved techniques for training gans." Advances in neural information processing systems 29 (2016).

[15] Devlin, Jacob, Ming-Wei Chang, Kenton Lee, and Kristina Toutanova. "Bert: Pre-training of deep bidirectional transformers for language understanding." arXiv preprint, arXiv:1810.04805 (2018).

[16] Clark, Kevin, Minh-Thang Luong, Quoc V. Le, and Christopher D. Manning. "Electra: Pre-training text encoders as discriminators rather than generators." arXiv preprint, arXiv:2003.10555 (2020).

[17] Bhattacharjee, Abhik, Tahmid Hasan, Wasi Ahmad Uddin, Kazi Mubasshir, Md Saiful Islam, Anindya Iqbal, M. Sohel Rahman, and Rifat Shahriyar. "Banglabert: Lagnuage model pretraining and benchmarks for low-resource language understanding evaluation in bangla." Findings of the North American Chapter of the Association for Computational Linguistics: NAACL (2022).

[18] Lan, Zhenzhong, Mingda Chen, Sebastian Goodman, Kevin Gimpel, Piyush Sharma, and Radu Soricut. "Albert: A lite bert for self-supervised learning of language representations." arXiv preprint, arXiv:1909.11942 (2019).

[19] Hasan, Tahmid, Abhik Bhattacharjee, Kazi Samin, Masum Hasan, Madhusudan Basak, M. Sohel Rahman, and Rifat Shahriyar. "Not low-resource anymore: Aligner ensembling, batch filtering, and new datasets for Bengali-English machine translation." arXiv preprint, arXiv:2009.09359 (2020).